\documentclass[sw]{iosart2x}

\usepackage{booktabs} 
\usepackage{makeidx}  
\usepackage{graphicx}
\usepackage{hyperref}
\usepackage{listings}
\usepackage{amsmath}
\usepackage{multicol}
\usepackage{amsthm}
\usepackage[linesnumbered]{algorithm2e}
\SetKwProg{Fn}{Function}{}{}

\usepackage{xspace}

\usepackage{todonotes} 

\newtheorem{definition}{Definition}

\newcommand{\myurl}[1]{{\fontsize{8}{8}\selectfont{\url{#1}}}}

\newcommand{\todoiteminline}[3]{
    \todoitemtemplate{#1}{#2}{#3}{inline}{red}
}

\newcommand{\todoitemtemplate}[5]{%
\todo[#4,color=#5,caption=X]{{#1}{ \textbf{{\tiny{for}} #2}:}{#3}}%
}

\newcommand{\code}[1]{\texttt{#1}}

\newcommand{\ie}{i.e.~}

\newcommand{\eg}{e.g.,~}

\newcommand{\qq}[1]{``#1''}

\newcommand{\NAME}{WDAqua-core1\xspace}

\usepackage{tabulary}
\newcolumntype{K}[1]{>{\centering\arraybackslash}p{#1}}

\pubyear{0000}
\volume{0}
\firstpage{1}
\lastpage{1}

\begin{document}
\begin{frontmatter}
\title{Towards a Question Answering System over the Semantic Web}
\runningtitle{Towards a Question Answering System over the Semantic Web}

\author[A]{\inits{N.}\fnms{Dennis} \snm{Diefenbach}\ead[label=e1]{dennis.diefenbach@univ-st-etienne.fr}},
\author[B]{\inits{N.}\fnms{Andreas} \snm{Both}\ead[label=e1]{contact@andreasboth.de}},
\author[A]{\inits{N.}\fnms{Kamal} \snm{Singh}\ead[label=e1]{kamal.singh@univ-st-etienne.fr}}
and
\author[A]{\inits{N.}\fnms{Pierre} \snm{Maret}\ead[label=e1]{pierre.maret@univ-st-etienne.fr}}
\runningauthor{D. Diefenbach et al.}
\address[A]{\orgname{Universit\'{e} de Lyon, CNRS UMR 5516 Laboratoire Hubert Curien}, \cny{France}}
\address[B]{\orgname{DATEV eG}, \cny{Germany}}





\begin{abstract}
Thanks to the development of the Semantic Web, a lot of new structured data has become available on the Web in the form of knowledge bases (KBs). Making this valuable data accessible and usable for end-users is one of the main goals of Question Answering (QA) over KBs. Most current QA systems query one KB, in one language (namely English). The existing approaches are not designed to be easily adaptable to new KBs and languages.\\
We first introduce a new approach for translating natural language questions to SPARQL queries. It is able to query several KBs simultaneously, in different languages, and can easily be ported to other KBs and languages. In our evaluation, the impact of our approach is proven using 5 different well-known and large KBs: Wikidata, DBpedia, MusicBrainz, DBLP and Freebase as well as 5 different languages namely English, German, French, Italian and Spanish. Second, we show how we integrated our approach, to make it easily accessible by the research community and by end-users.\\
To summarize, we provided a conceptional solution for multilingual, KB-agnostic Question Answering over the Semantic Web. The provided first approximation validates this concept.
\end{abstract}

\begin{keyword}
\kwd{Question Answering}
\kwd{Multilinguality}
\kwd{Portability}
\kwd{QALD}
\kwd{SimpleQuestions}
\end{keyword}

\end{frontmatter}

\section{Introduction}
Question Answering (QA) is an old research field in computer science that started in the sixties~\cite{lopez2011question}.
In the Semantic Web, a lot of new structured data has become available in the form of knowledge bases (KBs). 
Nowadays, there are KBs about media, publications, geography, life-science and more\footnote{\url{http://lod-cloud.net}}. 
The core purpose of a QA system over KBs is to retrieve the desired information from one or many KBs, using natural language questions. 
This is generally addressed by translating a natural language question to a SPARQL query.
Current research does not address the challenge of multilingual, KB-agnostic QA for both full and keyword questions (Table~\ref{tab:SoAQA}).

There are multiple reasons for that. 
Many QA approaches rely on language-specific tools (NLP tools), \eg SemGraphQA~\cite{beaumont2015semgraphqa}, gAnswer~\cite{zou2014natural} and Xser~\cite{xu2014xser}. 
Therefore, it is difficult or impossible to port them to a language-agnostic system. 
Additionally, many approaches make particular assumptions on how the knowledge is modelled in a given KB (generally referred to as \qq{structural gap}~\cite{diefenbach2017core}). 
This is the case of AskNow~\cite{dubey2016asknow} and DEANNA~\cite{yahya2012natural}.\\ 
There are also approaches which are difficult to port to new languages or KBs because they need a lot of training data which is difficult and expensive to create. 
This is for example the case of Bordes et al.~\cite{bordes2015large}. 
Finally there are approaches where it was not proven that they scale well. 
This is for example the case of SINA~\cite{shekarpour2015sina}.\\
In this paper, we present an algorithm that addresses all of the above drawbacks and that can compete, in terms of F-measure, with many existing approaches.\\
This publication is organized as follows. 
In section \ref{related_work}, we present related works. 
In section \ref{approach} and \ref{implementation}, we describe the algorithm providing the foundations of our approach. 
In section \ref{eval}, we provide the results of our evaluation over different benchmarks. 
In section \ref{qanary}, we show how we implemented our algorithm as a service so that it is easily accessible to the research community, and how we extended a series of existing services so that our approach can be directly used by end-users. 
We conclude with section \ref{conclusion}.

\begin{table}[]
    \centering
    \begin{tabular}{|K{2.7cm}|K{0.9cm}|K{1.5cm}|K{0.7cm}|}
        \hline
         \textbf{QA system} & \textbf{Lang} & \textbf{KBs} & \textbf{Type}\\
         \hline
         gAnswer~\cite{zou2014natural} (QALD-3 Winner)& en & DBpedia & full\\
         \hline
         Xser~\cite{xu2014xser} (QALD-4 \& 5 Winner)& en & DBpedia & full\\
         \hline
         UTQA~\cite{UTQA}& en, es, fs & DBpedia & full \\
         \hline
          Jain~\cite{jain2016question} (WebQuestions Winner)& en & Freebase & full \\
          \hline
         Lukovnikov~\cite{lukovnikov2017neural} (SimpleQuestions Winner)& en & Freebase & full\\
         \hline
         Ask Platypus (\url{https://askplatyp.us}) & en & Wikidata & full\\
         \hline
         \NAME  & en, fr, de, \newline it, es & Wikidata, DBpedia, Freebase, DBLP, MusicBrainz & full \& key\\
         \hline
    \end{tabular}
    \caption{Selection of QA systems evaluated over the most popular benchmarks. We indicated their capabilities with respect to multilingual questions, different KBs and different typologies of questions (full = \qq{well-formulated natural language questions}, key = \qq{keyword questions}).
    }
    \label{tab:SoAQA}
    \vspace{-1mm}
\end{table}

\section{Related work}\label{related_work}
In the context of QA, a large number of systems have been developed in the last years. 
For a complete overview, we refer to \cite{diefenbach2017core}. 
Most of them were evaluated on one of the following three popular benchmarks: WebQuestions~\cite{berant2013semantic}, SimpleQuestions~\cite{bordes2015large} and QALD\footnote{
\url{http://www.sc.cit-ec.uni-bielefeld.de/qald/}}.\\ 
WebQuestions contains 5810 questions that can be answered by one reefied statement. SimpleQuestions contains 108442 questions that can be answered using a single, binary-relation. The QALD challenge versions include more complex questions than the previous ones, and contain between 100 and 450 questions, and are therefore, compared to the other, small datasets.\\
The high number of questions of WebQuestions and SimpleQuestions led to many supervised-learning approaches for QA. Especially deep learning approaches became very popular in the recent years like Bordes et al.~\cite{bordes2015large} and Zhang et al.~\cite{zhang2016question}. 
The main drawback of these approaches is the training data itself. 
Creating a new training dataset for a new language or a new KB might be very expensive. 
For example, Berant et al.~\cite{berant2013semantic}, report that they spent several thousands of dollars for the creation of WebQuestions using Amazon Mechanical Turk. 
The problem of adapting these approaches to new dataset and languages can also be seen by the fact that all these systems work only for English questions over Freebase.\\ 
A list of the QA systems that were evaluated with QALD-3, QALD-4, QALD-5, QALD-6 can be found in Table~\ref{tab:QALD Benchmark}.
According to \cite{diefenbach2017core} less than 10\% of the approaches were applied to more than one language and 5\% to more than one KB. 
The reason is the heavy use of NLP tools or NL features like in Xser~\cite{xu2014xser}, gAnswer~\cite{zou2014natural} or QuerioDali~\cite{lopez2016queriodali}.\\
The problem of QA in English over MusicBrainz\footnote{\url{https://musicbrainz.org}} was proposed in QALD-1, in the year 2011. 
Two QA systems tackled this problem. 
Since then the MusicBrainz KB\footnote{\url{https://github.com/LinkedBrainz/MusicBrainz-R2RML}} completely changed. 
We are not aware of any QA system over DBLP\footnote{\url{http://dblp.uni-trier.de}}.\\ 
In summary, most QA systems work only in English and over one KB. 
Multilinguality is poorly addressed while portability is generally not addressed at all.\\
The fact that QA systems often reuse existing techniques and need several services to be exposed to the end-user, leads to the idea of developing QA systems in a modular way. 
At least four frameworks tried to achieve this goal: QALL-ME~\cite{DBLP:journals/ws/FerrandezSKDFNITONMG11}, openQA~\cite{Marx:2014:TOQ:2660517.2660519}, the Open Knowledge Base and Question-Answering (OKBQA) challenge\footnote{\url{http://www.okbqa.org/}} and Qanary \cite{2016:ICSC:QAVocabulary,bothqanary,dennisqanary}. 
We integrated our system as a Qanary QA component called \NAME. 
We choose Qanary for two reasons. 
First, it offers a series of off-the-shelf services related to QA systems and second, it allows to freely configure a QA system based on existing QA components.

\section{Approach for QA over Knowledge Bases}\label{approach}

\begin{figure}
\begin{center}
\includegraphics[width=0.2\textwidth]{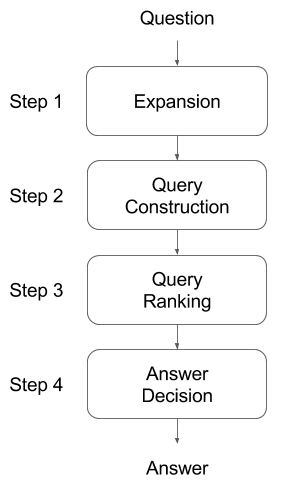}
\caption{Conceptual overview of the approach}\label{fig:conceptual} 
\end{center}
\vspace{-5mm}
\end{figure}

In this section, we present our multilingual, KB-agnostic approach for QA. 
It is based on the observation that many questions can be understood from the semantics of the words in the question while the syntax of the question has less importance.
For example, consider the question \qq{Give me actors born in Berlin}. 
This question can be reformulated in many ways like \qq{In Berlin were born which actors?}\ or as a keyword question \qq{Berlin, actors, born in}. 
In this case by knowing the semantics of the words \qq{Berlin}, \qq{actors}, \qq{born}, we are able to deduce the intention of the user. 
This holds for many questions, \ie they can be correctly interpreted without considering the syntax as the semantics of the words is sufficient for them. 
Taking advantage of this observation is the main idea of our approach. 
The KB encodes the semantics of the words and it can tell what is the most probable interpretation of the question (w.r.t.~the knowledge model described by the KB).\\
Our approach is decomposed in 4 steps: question expansion, query construction, query ranking and response decision. 
A conceptual overview is given in Figure~\ref{fig:conceptual}. 
In the following, the processing steps are described. 
As a running example, we consider the question \qq{Give me philosophers born in Saint-Etienne}. 
For the sake of simplicity, we use DBpedia as KB to answer the question.
However, it is important to recognize that no assumptions either about the language or the KB are made. 
Hence, even the processing of the running example is language- and KB-agnostic.

\begin{table*}
\scalebox{0.9}{
\begin{tabular}{|c|c|c|c|c|c|c|c|c|}
\hline
n & start & end & n-gram & resource \\
\hline
1 & 2 & 3 & philosophers & \code{dbrc:Philosophes}\\
2 & 2 & 3 & philosophers & \code{dbr:Philosophes} \\
3 & 2 & 3 & philosophers & \code{dbo:Philosopher} \\
4 & 2 & 3 & philosophers & \code{dbrc:Philosophers} \\
5 & 2 & 3 & philosophers & \code{dbr:Philosopher} \\
6 & 2 & 3 & philosophers & \code{dbr:Philosophy}  \\
7 & 2 & 3 & philosophers & \code{dbo:philosophicalSchool} \\
8 & 3 & 4 & born & \code{dbr:Born,\_Netherlands} \\
9 & 3 & 4 & born & \code{dbr:Born\_(crater)} \\
10 & 3 & 4 & born & \code{dbr:Born\_auf\_dem\_Dar?} \\
11 & 3 & 4 & born & \code{dbr:Born,\_Saxony-Anhalt} \\
\hline
\vdots\\
\hline
42 & 3 & 4 & born & \code{dbp:bornAs} \\
43 & 3 & 4 & born & \code{dbo:birthDate} \\
44 & 3 & 4 & born & \code{dbo:birthName} \\
45 & 3 & 4 & born & \code{dbp:bornDay} \\
46 & 3 & 4 & born & \code{dbp:bornYear} \\
47 & 3 & 4 & born & \code{dbp:bornDate} \\
48 & 3 & 5 & born in & \code{dbp:bornIn} \\
49 & 3 & 5 & born in & \code{dbo:birthPlace} \\
50 & 3 & 5 & born in & \code{dbo:hometown} \\
\hline
\end{tabular}

\begin{tabular}{|c|c|c|c|c|c|c|c|c|}
\hline
n & start & end & n-gram & resource \\
\hline
52 & 5 & 6 & saint & \code{dbr:SAINT\_(software)} \\
53 & 5 & 6 & saint & \code{dbr:Saint} \\
54 & 5 & 6 & saint & \code{dbr:Boxers\_and\_Saints} \\ 
55 & 5 & 6 & saint & \code{dbr:Utah\_Saints} \\
56 & 5 & 6 & saint & \code{dbr:Saints,\_Luton} \\
57 & 5 & 6 & saint & \code{dbr:Baba\_Brooks} \\
58 & 5 & 6 & saint & \code{dbr:Battle\_of\_the\_Saintes} \\
59 & 5 & 6 & saint & \code{dbr:New\_York\_Saints} \\
\hline
\vdots\\
\hline
106 & 5 & 6 & saint & \code{dbp:saintPatron} \\
107 & 5 & 6 & saint & \code{dbp:saintsDraft} \\
108 & 5 & 6 & saint & \code{dbp:saintsSince} \\
109 & 5 & 6 & saint & \code{dbo:patronSaint} \\
110 & 5 & 6 & saint & \code{dbp:saintsCollege} \\
111 & 5 & 6 & saint & \code{dbp:patronSaintOf} \\
112 & 5 & 6 & saint & \code{dbp:patronSaint(s)} \\
113 & 5 & 6 & saint & \code{dbp:patronSaint'sDay} \\
114 & 5 & 7 & saint etienne & \code{dbr:Saint\_Etienne\_(band)} \\
115 & 5 & 7 & saint etienne & \code{dbr:Saint\_Etienne} \\
116 & 5 & 7 & saint etienne & \code{dbr:Saint-\'Etienne} \\
117 & 6 & 7 & etienne & \code{dbr:\'Etienne} \\
\hline
\end{tabular}
}
\caption{
Expansion step for the question \qq{Give me philosophers born in Saint \'Etienne}. 
The first column enumerates the candidates that were found. 
Here, 117 possible entities, properties and classes were found from the question. 
The second, third and fourth columns indicate the position of the n-gram in the question and the n-gram itself. 
The last column is for the associated IRI.
Note that many possible meanings are considered: line 9 says that \qq{born}\ may refer to a crater, line 52 that \qq{saint}\ may refer to a software and line 114 that the string \qq{Saint \'Etienne}\ may refer to a band.}
\label{exp}
\end{table*}

\subsection{Expansion}\label{expansion}
Following a recent survey \cite{diefenbach2017core}, we call a lexicalization, a name of an entity, a property or a class. 
For example, \qq{first man on the moon}\ and \qq{Neil Armstrong}\ are both lexicalizations of \code{dbr:Neil\_Armstrong}. 
In this step, we want to identify all entities, properties and classes, which the question could refer to. 
To achieve this, we use the following rules:
\begin{itemize}
\item All IRIs are searched whose lexicalization (up to stemming) is an n-gram $N$ (up to stemming) in the question.
\item If an n-gram $N$ is a stop word (like ``is'', ``are'', ``of'', ``give'', \ldots), then we exclude the IRIs associated to it. 
This is due to the observation that the semantics are important to understand a question and the fact that stop words do not carry a lot of semantics.
Moreover, by removing the stop words the time needed in the next step is decreased.
\end{itemize}
An example is given in Table \ref{exp}. 
The stop words and the lexicalizations used for the different languages and KBs are described in section~\ref{lexicalizations}. 
In this part, we used a well-known Apache Lucene Index\footnote{https://lucene.apache.org} technology which allows fast retrieval, while providing a small disk and memory footprint.

\subsection{Query construction}\label{construction}
In this step, we construct a set of queries that represent possible interpretations of the given question within the given KB. 
Therefore, we heavily utilize the semantics encoded into the particular KB.
We start with a set $R$ of IRIs from the previous step. 
The goal is to construct all possible queries containing the IRIs in $R$ which give a non-empty result-set. 
Let $V$ be the set of variables. 
Based on the complexity of the questions in current benchmarks, we restrict our approach to queries satisfying 4 patterns:\\
\begin{align*}
&\mbox{\code{SELECT / ASK var}}\\
&\mbox{\code{ WHERE \{  s1 s2 s3 . \}}}\\
\end{align*}
\begin{align*}
&\mbox{\code{SELECT / ASK var}} \\
&\mbox{\code{WHERE \{ s1 s2 s3 .}}\\
&\mbox{\quad\quad\quad\quad\code{  s4 s5 s6 . \}}}
\end{align*}\\
with
\begin{center}
\code{ s1, \ldots, s6} $\in \text{\code{R}} \cup \text{\code{V}}$\\
\end{center}
and
\begin{center}
$\text{\code{var}} \in \text{\code{\{s1, \ldots ,s6\}}} \cap \text{\code{V}}$\\
\end{center}
, i.e. all queries containing one or two triple patterns that can be created starting from the IRIs in $R$. 
Moreover, for entity linking, we add the following two patterns:\\
\begin{align*}
&\mbox{\code{SELECT ?x}}\\ 
&\mbox{\code{WHERE \{ VALUES { ?x \{iri\}} . \}}}
\end{align*}
\\
\begin{align*}
&\mbox{\code{SELECT ?x}}\\ 
&\mbox{\code{WHERE \{ VALUES { ?x \{iri\}} . }}\\
&\mbox{\quad\code{iri ?p iri1 . \}}}
\end{align*}\\
\code{with iri, iri1} $\in$ \code{R}, i.e. all queries returning directly one of the IRIs in \code{R} with possibly one additional triple.\\
Note that these last queries just give back directly an entity and should be generated for a question like: \qq{What is Apple Company?}\ or \qq{Who is Marie Curie?}. 
An example of generated queries is given in Figure~\ref{construct}.\\
The main challenge is the efficient construction of these SPARQL queries. 
The main idea is to perform in the KB graph a breadth-first search of depth $2$ starting from every IRI in $R$. 
While exploring the KB for all IRIs $r_j \in R$ (where $r_j \neq r_i$) the distance $d_{r_i,r_j}$ between two resources is stored. 
These numbers are used when constructing the queries shown above. 
For a detailed algorithm of the query construction phase, please see section~\ref{implementation}. 
Concluding, in this section, we computed a set of possible SPARQL queries (candidates). 
They are driven by the lexicalizations computed in section~\ref{expansion}\ and represent the possible intentions expressed by the question of the user.

\begin{figure}[!t]
\begin{itemize}
\item 
\begin{lstlisting}
SELECT DISTINCT ?y WHERE { 
  dbr:Saint_(song) ?p ?x . 
  ?x  dbo:hometown  ?y . }
\end{lstlisting}\leavevmode
\item 
\begin{lstlisting}
SELECT ?x { 
VALUES ?x { dbr:Saint_Etienne_(band) } } 
\end{lstlisting}\leavevmode
\item 
\begin{lstlisting}
SELECT DISTINCT ?y WHERE { 
  ?x dbo:birthPlace dbr:Saint-Etienne . 
  ?x dbo:birthDate  ?y . }
\end{lstlisting}\leavevmode
\item 
\begin{lstlisting}
SELECT DISTINCT ?y WHERE { 
  ?x ?p dbr:Philosophy  . 
  ?x dbo:birthDate  ?y . }
\end{lstlisting}\leavevmode
\end{itemize}
    \caption{Some of the 395 queries constructed for the question \qq{Give me philosophers born in Saint Etienne.}. Note that all queries could be semantically related to the question. The second one is returning \qq{Saint-Etienne}\ as a band, the third one the birth date of people born in the city of \qq{Saint-Etienne}\ and the forth one the birth date of persons related to philosophy.}
\label{construct}
\end{figure}
 
\subsection{Ranking}\label{ranking}
Now the computed candidates need to be ordered by their probability of answering the question correctly.
Hence, we rank them based on the following features:
\begin{itemize}
    \item Number of the words in the question which are covered by the query. For example, the first query in Figure \ref{construct} is covering two words (``Saint'' and ``born'').
    \item The edit distance of the label of the resource and the word it is associated to. For example, the edit distance between the label of \code{dbp:bornYear}\ (which is ``born year'') and the word ``born'' is $5$.
    \item The sum of the relevance of the resources, (e.g. the number of inlinks and the number of outlinks of a resource). This is a knowledge base independent choice, but it is also possible to use a specific score for a KB (like page-rank).
    \item The number of variables in the query.
    \item The number of triples in the query.
\end{itemize} 
If no training data is available, then we rank the queries using a linear combination of the above 5 features, where the weights are determined manually.
Otherwise we assume a training dataset of questions together with the corresponding answers set, which can be used to calculate the F-measure for each of the SPARQL query candidates. 
As a ranking objective, we want to order the SPARQL query candidates in descending order with respect to the F-measure.
In our exemplary implementation we rank the queries using RankLib\footnote{\url{https://sourceforge.net/p/lemur/wiki/RankLib/}} with Coordinate Ascent~\cite{metzler2007linear}. 
At test time the learned model is used to rank the queries, the top-ranked query is executed against a SPARQL endpoint, and the result is computed. 
An example is given in Figure~\ref{rank}. 
Note that, we do not use syntactic features. 
However, it is possible to use them to further improve the ranking.
\subsection{Answer Decision}\label{answerdecision}
\newcommand{\thresholdA}{\ensuremath{\theta_1}}
\newcommand{\thresholdB}{\ensuremath{\theta_2}}
The computations in the previous section lead to a list of ranked SPARQL queries candidates representing our possible interpretations of the user's intentions.
Although the quality of this processing step is high (as shown in several experiments), an additional confidence score is computed.
We construct a model based on logistic regression. 
We use a training set consisting of SPARQL queries and the labels true or false. 
True indicates if the F-score of the SPARQL query is bigger than a threshold \thresholdA\ or false otherwise. 
Once the model is trained, it can compute a confidence score $p_Q \in [0,1]$ for a query $Q$.
In our exemplary implementation we assume a correctly ordered list of SPARQL query candidates computed in section~\ref{ranking}.
Hence, it only needs to be checked whether $p_{Q_1} \geq \thresholdB$ is true for the first ranked query $Q_1$ of the SPARQL query candidates, or otherwise it is assumed that the whole candidate list is not reflecting the user's intention.
Hence, we refuse to answer the question.
We answer the question if it is above a threshold \thresholdB\, otherwise we do not answer it. 
Note that $p_Q$ can be interpreted as the confidence that the QA system has in the generated SPARQL query $Q$, i.e.~in the generated answer. 

\begin{figure}[t!]
\begin{enumerate}
\item 
\begin{lstlisting}
SELECT DISTINCT ?x WHERE { 
  ?x dbp:birthPlace dbr:Saint-Etienne . 
  ?x rdf:type dbo:Philosopher . }
\end{lstlisting}\leavevmode
\item \begin{lstlisting}
SELECT DISTINCT ?y WHERE { 
  ?x  dbo:birthPlace dbr:Saint-Etienne . 
  ?x  dbo:philosophicalSchool ?y . } 
\end{lstlisting}\leavevmode
\item 
\begin{lstlisting}
SELECT DISTINCT ?x WHERE { 
 ?x  dbp:birthPlace dbr:Saint-Etienne . }
\end{lstlisting}\leavevmode
\item
\begin{lstlisting}
SELECT DISTINCT ?x WHERE { 
 ?x  dbo:hometown dbr:Saint-Etienne . }
\end{lstlisting}
 \end{enumerate}
\caption{The top 4 generated queries for the question ``Give me philosophers born in Saint \'Etienne.''. (1) is the query that best matches the question; (2) gives philosophical schools of people born in Saint-\'Etienne; (3)(4) give people born in Saint-\'Etienne or that live in Saint-\'Etienne. The order can be seen as a decreasing approximation to what was asked.}
\label{rank}
\end{figure}

\subsection{Multiple KBs}

Note that the approach can also be extended, as it is, to multiple KBs. 
In the query expansion step, one has just to take in consideration the labels of all KBs. 
In the query construction step, one can consider multiple KBs as one graph having multiple unconnected components. 
The query ranking and answer decision step are literally the same.

\subsection{Discussion}
Overall, we follow a combinatorial approach with efficient pruning, that relies on the semantics encoded in the underlying KB.\\
In the following, we want to emphasize the advantages of this approach using some examples.
\begin{itemize}
\item \textbf{Joint disambiguation of entities and relations:}
For example, for interpreting the question ``How many inhabitants has Paris?'' between the hundreds of different meanings of ``Paris'' and ``inhabitants'' the top ranked queries contain the resources called ``Paris'' which are cities, and the property indicating the population, because only these make sense semantically.
\item \textbf{Portability to different KBs:} One problem in QA over KBs is the semantic gap, 
i.e.~the difference between how we think that the knowledge is encoded in the KB and how it actually is. 
For example, in our approach, for the question ``What is the capital of France?'', we generate the query
\begin{lstlisting}
SELECT ?x WHERE { 
  dbr:France dbp:capital ?x . 
}
\end{lstlisting}
which probably most users would have expected, but also the query
\begin{lstlisting}
SELECT ?x { 
  VALUES ?x {
    dbr:List_of_capitals_of_France
  } 
}
\end{lstlisting}
which refers to an overview article in Wikipedia about the capitals of France and that most of the users would probably not expect. 
This important feature allows to port the approach to different KBs while it is independent of how the knowledge is encoded.
\item \textbf{Ability to bridge over implicit relations:} We are able to bridge over implicit relations. For example, given ``Give me German mathematicians'' the following query is computed:
\begin{lstlisting}
SELECT DISTINCT ?x WHERE { 
  ?x  ?p1 dbr:Mathematician . 
  ?x  ?p2 dbr:Germany . 
}
\end{lstlisting}
Here \code{?p1} is:
\begin{itemize}[$\bullet$]
    \item \code{dbo:field}
    \item \code{dbo:occupation},
    \item \code{dbo:profession}
\end{itemize}
and \code{?p2} is:
\begin{itemize}[$\bullet$]
    \item \code{dbo:nationality},
    \item \code{dbo:birthPlace},
    \item \code{dbo:deathPlace},
    \item  \code{dbo:residence}.
\end{itemize}
Note that all these properties could be intended for the given question.
\item \textbf{Easy to port to new languages:} The only parts where the language is relevant are the stop word removal and stemming. Since these are very easy to adapt to new languages, one can port the approach easily to other languages.
\item  \textbf{Permanent system refinement:} It is possible to improve the system over time. The system generates multiple queries. This fact can be used to easily create new training dataset as it is shown in \cite{dennisBenchmark}. Using these datasets one can refine the ranker to perform better on the asked questions.
\item \textbf{System robust to malformed questions and keyword questions:} There are no NLP tools used in the approach which makes it very robust to malformed questions. For this reason, keyword questions are also supported.
\end{itemize}
A disadvantage of our exemplary implementation is that the identification of relations relies on a dictionary. 
Note that, non-dictionary based methods follow one of the following strategies. 
Either they try to learn ways to express the relation from big training corpora (like in~\cite{bordes2015large}), s.t.~the problem is shifted to create suitable training sets. 
Or text corpora are used to either extract lexicalizations for properties (like in~\cite{berant2013semantic}) or learn word-embeddings (like in \cite{hakimov2017amuse}).
Hence, possible improvements might be applied to this task in the future.
\section{Fast candidate generation}\label{implementation}
In this section, we explain how the SPARQL queries described in section \ref{construction} can be constructed efficiently. 

Let $R$ be a set of resources.
We consider the KB as a directed labeled graph $G$:
\begin{definition}(Graph)
A directed labeled graph is an ordered pair $G=(V,E,f)$, such that:
\begin{itemize}
\item $V$ is a non-empty set, called the vertex set;
\item $E$ is a set, called edge set, such that $E\subset \{(v,w): v,w\in V\}$, i.e.~a subset of the pairs of $V$;
\item For a set $L$ called labeled set, $f$ is a function $f:E\rightarrow L$, i.e.~a function that assigns to each edge a label $p\in L$. 
We indicate an edge with label $p$ as $e=(v,p,w)$.
\end{itemize}
\end{definition}

To compute the pairwise distance in $G$ between every resource in $R$, we do a breadth-first search from every resource in $R$ in an undirected way (i.e. we traverse the graph in both directions).\\
We define a distance function $d$ as follows. Assume we start from a vertex $r$ and find the following two edges $e_{1}=(r,p_{1},r_{1})$, $e_{2}=(r_{1},p_{2},r_{2})$. 
We say that $d_{r,p_{1}} = 1$, $d_{r,r_{1}} = 2$, $d_{r,p_{2}} = 3$ and so on. When an edge is traversed in the opposite direction, we add a minus sign. 
For example, given the edges $e_{1}=(r,p_{1},r_{1})$ and $e_{2}=(r_{2},p_{2},r_{1})$, we say $d_{r,p_{2}} = -3$. For a vertex or edge $r$, and a variable $x$ we artificially set $d_{r,x}$ to be any possible integer number. Moreover, we set $ d_{x,y}=d_{y,x}$ for any $x,y$. The algorithm to compute these numbers can be found in Algorithm~\ref{fig:algo1}. 

\begin{algorithm}
 \KwData{Graph $G=(V,E,f)$ and a set $R$ of edges and labels}
 \KwResult{The pairwise distance between elements in $R$}
 \For{$r\in R\cap V$}{
    \For{$e_{1}$=($r$,$p_{1}$,$r_{1}$)$\in$ E}{
        \textbf{if} $p_{1}\in R$ \textbf{then} $d_{r,p_{1}}=1$; \textbf{if} $r_{1}\in R$ \textbf{then} $d_{r,l_{1}}=2$\\
        \For{$(e_{2}=(r_{1},p_{2},r_{2})\in E)$}{
            \textbf{if} $p_{2}\in R$ \textbf{then} $d_{r,p_{2}}=3$; \textbf{if} $r_{2}\in R$ \textbf{then} $d_{r,2_{2}}=4$;\\ \textbf{if} $p_{1},p_{2}\in R$ \textbf{then} $d_{p_{1},p_{2}}=2$: \textbf{if} $p_{1},r_{2}\in R$ \textbf{then} $d_{p_{1},r_{2}}=3$\\
        }
        \For{$(e_{2}=(r_{2},p_{2},r_{1})\in E)$}{
            \textbf{if} $p_{2}\in R$ \textbf{then} $d_{r,p_{2}}=-3$; \textbf{if} $r_{2}\in R$ \textbf{then} $d_{r,2_{2}}=-4$\\
            \textbf{if} $p_{1},p_{2}\in R$ \textbf{then} $d_{p_{1},p_{2}}=-2$; \textbf{if} $p_{1},p_{2}\in R$ \textbf{then} $d_{p_{1},2_{2}}=-3$\\
        }
    }
    \For{$e_{1}$=($r_{1}$,$p_{1}$,$r$)$\in E$}{
        \textbf{if} $p_{1}\in R$ \textbf{then} $d_{r,p_{1}}=-1$; \textbf{if} $r_{1}\in R$ \textbf{then} $d_{r,l_{1}}=-2$\\
        \For{$(e_{2}=(r_{1},p_{2},r_{2})\in E)$}{
            \textbf{if} $p_{2}\in R$ \textbf{then} $d_{r,p_{2}}=3$; \textbf{if} $r_{2}\in R$ \textbf{then} $d_{r,2_{2}}=4$\\
            \textbf{if} $p_{1},p_{2}\in R$ \textbf{then} $d_{p_{1},p_{2}}=2$; \textbf{if} $p_{1},r_{2}\in R$ \textbf{then} $d_{p_{1},r_{2}}=3$\\
        }
        \For{$(e_{2}=(r_{2},p_{2},r_{1})\in E)$}{
            \textbf{if} $p_{2}\in R$ \textbf{then} $d_{r,p_{2}}=3$; \textbf{if} $r_{2}\in R$ \textbf{then} $d_{r,2_{2}}=4$\\
            \textbf{if} $p_{1},p_{2}\in R$ \textbf{then} $d_{p_{1},p_{2}}=2$; \textbf{if} $p_{1},p_{2}\in R$ \textbf{then} $d_{p_{1},2_{2}}=3$\\
        }
    }
 }
\caption{Algorithm to compute the pairwise distance between every resource in a set $R$ appearing in a KB.}
\label{fig:algo1}
\end{algorithm}

\begin{algorithm*}
\KwData{Graph $G=(V,E,f)$ and a set $R$ of vertices and edges, and their pairwise distance $d$}
\KwResult{All connected triple patterns in $G$ from a set $R$ of vertices and edges with maximal $K$ triple patterns}
$L=\emptyset$  \#list of triple patterns\\
$V_{s,o}=\emptyset$  \#set of variables in subject, object position\\
$V_{p}=\emptyset$  \#set of variables in predicate position\\
k=0\\
\Fn{generate (L,k)}{
    \For{$s_{1} \in (R\cap V)\cup V_{s,o} \cup \{x_{k,1}\}$}{
        \For{$s_{2} \in (R\cap P)\cup V_{p} \cup \{x_{k,2}\}$}{
            \For{$s_{3} \in (R\cap V)\cup V_{s,o} \cup \{x_{k,3}\}$}{
                \textbf{if} $k=0 \wedge d_{s_{2},s_{3}}=-1 \wedge d_{s_{1},s_{2}}=1  \wedge d_{s_{1},s_{3}}=2$ \textbf{then} $L \leftarrow L \cup \{(s_{1},s_{2},s_{3})\}$\\
                \For{$T\in L^{(k)}$}{
                    $b_{1} = true; b_{2} = true; b_{3} = true; b_{4} = true;$\\
                    \For{$(t_{1},t_{2},t_{3}) \in T$}{
                        \textbf{if not} ($s_{1}=t_{1} \wedge d_{t_{1},s_{2}}=2 \wedge d_{t_{1},s_{3}}=3 \wedge d_{t_{2},s_{2}}=-2 \wedge d_{t_{2},s_{3}}=-3 \wedge d_{t_{3},s_{2}}=-3 \wedge d_{t_{3},s_{3}}=-4$) \textbf{then}  $b_{1} = false$ \\
                        \textbf{if not} ($s_{1}=t_{3} \wedge d_{t_{1},s_{2}}=3 \wedge d_{t_{1},s_{3}}=4 \wedge d_{t_{2},s_{2}}=2 \wedge d_{t_{2},s_{3}}=3 \wedge d_{t_{3},s_{2}}=1 \wedge d_{t_{3},s_{3}}=2)$ \textbf{then}  $b_{2} = false$ \\
                        \textbf{if not} ($s_{3}=t_{1} \wedge d_{t_{1},s_{2}}=-1 \wedge d_{t_{1},s_{3}}=-4 \wedge d_{t_{2},s_{2}}=-2 \wedge d_{t_{2},s_{3}}=-3 \wedge d_{t_{3},s_{2}}=-1 \wedge d_{t_{3},s_{3}}=-2$) \textbf{then}  $b_{3} = false$ \\
                       \textbf{if not} ($s_{3}=t_{3}\wedge d_{t_{1},s_{2}}=-3\wedge d_{t_{1},s_{3}}=2\wedge d_{t_{2},s_{2}}=-2\wedge d_{t_{2},s_{3}}=-1\wedge d_{t_{3},s_{2}}=-1$) \textbf{then}  $b_{4} = false$ \\
                    }
                    \If{ $b_{1}=true \vee b_{2}=true \vee b_{3}=true \vee b_{4}=true$}{ 
                    $ L\leftarrow L \cup (T \cup {(s_{1},s_{2},s_{3})})$;\\
                    $V_{s,o} \leftarrow  V_{s,o} \cup \{s_{1},s_{3}\}$;\\
                    $V_{p} \leftarrow  V \cup \{s_{2}\})$
                    }
                    \If{(k!=K)}{return \textit{generate(L,k+1)}}
                }
                
            }
        }
    }
}
\caption{Recursive algorithm to create all connected triple patterns from a set $R$ of resources with maximal $K$ triple patterns. 
$L$ contains the triple patterns created recursively and $L^{(k)}$ indicates the triple patterns with exactly $k$ triples. Note that the \qq{if not}\ conditions very often are not fulfilled. This guarantees the speed of the process.}
\label{fig:algo2}
\end{algorithm*}

The algorithm of our exemplary implementation simply traverses the graph starting from the nodes in $R$ in a breadth-first search manner and keeps track of the distances as defined above.
The breadth-first search is done by using HDT~\cite{fernandez2013binary} as an indexing structure\footnote{\url{https://www.w3.org/Submission/2011/03/}}. 
Note that HDT was originally developed as an exchange format for RDF files that is queryable. 
A rarely mentioned feature of HDT is that it is perfectly suitable for performing breadth-first search operations over RDF data. 
In HDT, the RDF graph is stored as an adjacency list which is an ideal data structure for breadth-first search operations. 
This is not the case for traditional triple-stores. 
The use of HDT at this point is key for two reasons, (1) the performance of the breadth-first search operations, and (2) the low footprint of the index in terms of disk and memory space. 
Roughly, a 100 GB RDF dump can be compressed to a HDT file of a size of approx.~10 GB~\cite{fernandez2013binary}.\\
Based on the numbers above, we now want to construct all triple patterns with $K$ triples and one projection variable recursively. 
Given a triple pattern $T$, we only want to build connected triple-pattern while adding triples to $T$. This can be done recursively using the algorithm described in Algorithm~\ref{fig:algo2}.
Note that thanks to the numbers collected during the breadth-first search operations, this can be performed very fast. 
Once the triple patterns are constructed, one can choose any of the variables, which are in subject or object position, as a projection variable.\\
The decision to generate a SELECT or and ASK query, is made depending on some regex expressions over the beginning of the question.

\section{Evaluation}\label{eval}
To validate the approach w.r.t. multilinguality, portability and robustness, we evaluated our approach using multiple benchmarks for QA that appeared in the last years. The different benchmarks are not comparable and they focus on different aspects of QA. For example SimpleQuestions focuses on questions that can be solved by one simple triple-pattern, while LC-QuAD focuses on more complex questions. 
Moreover, the QALD questions address different challenges including multilinguality and the use of keyword questions. 
Unlike previous works, we do not focus on one benchmark, but we analyze the behaviour of our approach under different scenarios. This is important, because it shows that our approach is not adapted to one particular benchmark, as it is often done by existing QA systems, and proofs its portability.\\
We tested our approach on 5 different datasets namely Wikidata\footnote{\url{https://www.wikidata.org/}}, DBpedia\footnote{\url{http://dbpedia.org}}, MusicBrainz\footnote{\url{https://musicbrainz.org}}, DBLP\footnote{\url{http://dblp.uni-trier.de}} and Freebase\footnote{\url{https://developers.google.com/freebase/}}. 
Moreover, we evaluated our approach on five different languages namely: English, German, French, Italian and Spanish. First, we describe how we selected stop words and collected lexicalizations for the different languages and KBs, then we describe and discuss our results.

\subsection{Stop Words and lexicalizations}\label{lexicalizations}
As stop words, we use the lists, for the different languages, provided by Lucene, together with some words which are very frequent in questions like \qq{what}, \qq{which}, \qq{give}.\\
Depending on the KB, we followed different strategies to collect lexicalizations. 
Since Wikidata has a rich number of lexicalizations, we simply took all lexicalizations associated to a resource through \code{rdfs:label}\footnote{\code{rdfs: http://www.w3.org/2000/01/rdf-schema\#}}, \code{skos:prefLabel}\footnote{\code{skos: http://www.w3.org/2004/02/skos/core\#}} and \code{skos:} \code{altLabel}. 
For DBpedia, we only used the English DBpedia, where first all lexicalizations associated to a resource through the \code{rdfs:label}\ property were collected. 
Secondly, we followed the disambiguation and redirect links to get additional ones and took also into account available demonyms \code{dbo:demonym} (i.e. to \code{dbr:Europe}\ we associate also the lexicalization \qq{European}).
Thirdly, by following the inter-language links, we associated the labels from the other languages to the resources. 
DBpedia properties are poorly covered with lexicalizations, especially when compared to Wikidata. 
For example, the property \code{dbo:birthPlace}\ has only one lexicalization namely \qq{birth place}, while the corresponding property over Wikidata \code{P19}\ has 10 English lexicalizations like \qq{birthplace}, \qq{born in}, \qq{location born}, \qq{birth city}. 
In our exemplary implementation two strategies were implemented.
First, while aiming at a QA system for the Semantic Web we also can take into account interlinkings between properties of distinguished KBs, s.t. lexicalizations are merged from all KBs currently considered.
There, the \code{owl:sameAs}\ links from DBpedia relations to Wikidata are used and every lexicalization present in Wikidata is associated to the corresponding DBpedia relation. 
Secondly, the DBpedia abstracts are used to find more lexicalizations for the relations.
To find new lexicalizations of a property $\textit{p}$ we follow the strategy proposed by \cite{gerber2011bootstrapping}. We extracted from the KB the subject-object pairs \textit{(x,y)} that are connected by \textit{p}. 
Then the abstracts are scanned and all sentences are retrieved which contain both $label(x)$ and $label(y)$. 
At the end, the segments of text between $label(x)$ and $label(y)$, or $label(y)$ and $label(x)$ are extracted. 
We rank the extracted text segments and we choose the most frequent ones. This was done only for English.\\
For MusicBrainz we used the lexicalizations attached to \code{purl:title}\footnote{purl: http://purl.org/dc/elements/1.1/}, \code{foaf:name}\footnote{foaf: http://xmlns.com/foaf/}, \code{skos:altLabel} and \code{rdfs:label}. 
For DBLP only the one attached to \code{rdfs:label}. 
Note, MusicBrainz and DBLP contain only few properties. 
We aligned them manually with Wikidata and moved the lexicalizations from one KB to the other. The mappings can be found under \url{http://goo.gl/ujbwFW} and \url{http://goo.gl/ftzegZ} respectively. 
This took in total 1 hour of manual work.\\
For Freebase, we considered the lexicalizations attached to \code{rdfs:label}. 
We also followed the few available links to Wikidata. Finally, we took the 20 most prominent properties in the training set of the SimpleQuestions benchmark and looked at the lexicalizations of them in the first 100 questions of SimpleQuestions. 
We extracted manually the lexicalizations for them. 
This took 1 hour of manual work. 
We did not use the other (75810 training and 10845 validation) questions, i.e. despite previews works we only took a small fraction of the available training data. 

\subsection{Experiments}
To show the performance of the approach on different scenarios, we benchmarked it using the following benchmarks.\\

\subsubsection{Benchmarks}
\textbf{QALD:} We evaluated our approach using the QALD benchmarks. 
These benchmarks are good to see the performance on multiple-languages and over both full-natural language questions and keyword questions. 
We followed the metrics of the original benchmarks. Note that the metrics changed in QALD-7. 
The results are given in Table~\ref{tab:QALD Benchmark} together with state-of-the-art systems. 
To find these, we used Google Scholar to select all publications about QA systems that cited one of the QALD challenge publications. 
Note that, in the past, QA systems were evaluated only on one or two of the QALD benchmarks.
We provide, for the first time, an estimation of the differences between the benchmark series. 
Over English, we outperformed 90\% of the proposed approaches. 
We do not beat Xser~\cite{xu2014xser} and UTQA~\cite{UTQA}. 
Note that these systems required additional training data than the one provided in the benchmark, which required a significant cost in terms of manual effort. 
Moreover, the robustness of these systems over keyword questions is probably not guaranteed.
We cannot prove this claim because for these systems neither the source code nor a web-service is available.\\
Due to the manual effort required to do an error analysis for all benchmarks and the limited space, we restricted to the QALD-6 benchmark. 
The error sources are the following. 
40\% are due to lexical gap (e.g. for ``Who played Gus Fring in Breaking Bad?'' the property \code{dbo:portrayer}\ is expected), 28\% come from wrong ranking, 12\% are due to the missing support of superlatives and comparatives in our implementation (e.g. ``Which Indian company has the most employees?''), 9\% from the need of complex queries with unions or filters (e.g. the question \qq{Give me a list of all critically endangered birds.} requires a filter on \code{dbo:conservationStatus}\ equal ``CR''), 6\% come from out of scope questions (i.e. question that should not be answered), 2\% from too ambiguous questions (e.g. ``Who developed Slack?'' is expected to refer to a ``cloud-based team collaboration tool'' while we interpret it as ``linux distribution''). 
One can see that keyword queries always perform worst as compared to full natural language queries. 
The reason is that the formulation of the keyword queries does not allow to decide if the query is an ASK query or if a COUNT is needed (e.g. ``Did Elvis Presley have children?'' is formulated as ``Elvis Presley, children''). 
This means that we automatically get these questions wrong.\\
To show the performance over Wikidata, we consider the QALD-7 task 4 training dataset. 
This originally provided only English questions. 
The QALD-7 task 4 training dataset reuses questions over DBpedia from previous challenges where translations in other languages were available. 
We moved these translations to the dataset. 
The results can be seen in Table~\ref{tab:wikidata_benchmark}. 
Except for English, keyword questions are easier than full natural language questions.
The reason is the formulation of the questions. 
For keyword questions the lexical gap is smaller. 
For example, the keyword question corresponding to the question ``Qui \'ecrivit Harry Potter?'' is ``\'ecrivain, Harry Potter''. 
Stemming does not suffice to map ``\'ecrivit'' to ``\'ecrivain'', lemmatization would be needed. 
This problem is much smaller for English, where the effect described over DBpedia dominates.
We can see that the best performing language is English, while the worst performing language is Italian. 
This is mostly related to the poorer number of lexicalizations for Italian. 
Note that the performance of the QA approach over Wikidata correlates with the number of lexicalizations for resources and properties for the different languages as described in \cite{kaffee2017glimpse}.
This indicates that the quality of the data, in different languages, directly affects the performance of the QA system. 
Hence, we can derive that our results will probably improve while the data quality is increased. 
Finally we outperform the presented QA system over this benchmark.\\

\begin{table}
\scalebox{0.78}{
\begin{tabular}{|c|c|c|c|c|c|c|c|c|}
\hline
QA system & Lang & Type & Total & P & R & F & Runtime & Ref\\
\hline
\multicolumn{9}{|c|}{QALD-3}\\
\hline

\textbf{\NAME} & \textbf{en} & \textbf{full} & \textbf{100} & \textbf{0.64} & \textbf{0.42} & \textbf{0.51} & \textbf{1.01} & -\\
\textbf{\NAME} & \textbf{en} & \textbf{key} & \textbf{100} & \textbf{0.71} & \textbf{0.37} & \textbf{0.48} & \textbf{0.79} & -\\
\textbf{\NAME} & \textbf{de} & \textbf{key} & \textbf{100} & \textbf{0.79} & \textbf{0.31} & \textbf{0.45} & \textbf{0.22} & -\\
\textbf{\NAME} & \textbf{de} & \textbf{full} & \textbf{100} & \textbf{0.79} & \textbf{0.28} & \textbf{0.42} & \textbf{0.30} & -\\
\textbf{\NAME} & \textbf{fr} & \textbf{key} & \textbf{100} & \textbf{0.83} & \textbf{0.27} & \textbf{0.41} & \textbf{0.26} & -\\
gAnswer~\cite{zou2014natural}$\ast$ & en & full & 100 & 0.40 & 0.40 & 0.40 & $\approx$ 1 s & \cite{zou2014natural}\\
\textbf{\NAME} & \textbf{fr} & \textbf{full} & \textbf{100} & \textbf{0.70} & \textbf{0.26} & \textbf{0.38} & \textbf{0.37} & -\\
\textbf{\NAME} & \textbf{es} & \textbf{full} & \textbf{100} & \textbf{0.77} & \textbf{0.24} & \textbf{0.37} & \textbf{0.27} & -\\
\textbf{\NAME} & \textbf{it} & \textbf{full} & \textbf{100} & \textbf{0.79} & \textbf{0.23} & \textbf{0.36} & \textbf{0.30} & -\\
\textbf{\NAME} & \textbf{it} & \textbf{key} & \textbf{100} & \textbf{0.84} & \textbf{0.23} & \textbf{0.36} & \textbf{0.24} & -\\
\textbf{\NAME} & \textbf{es} & \textbf{key} & \textbf{100} & \textbf{0.80} & \textbf{0.23} & \textbf{0.36} & \textbf{0.23} & -\\
RTV~\cite{giannone2013hmm} & en & full & 99 & 0.32 & 0.34 & 0.33 & - & \cite{cimiano2013multilingual}\\
Intui2~\cite{dima2013intui2} & en & full & 99 & 0.32 & 0.32 & 0.32 & - & \cite{cimiano2013multilingual}\\
SINA~\cite{shekarpour2015sina}$\ast$ & en & full & 100 & 0.32 & 0.32 & 0.32 & $\approx$ 10-20s & \cite{shekarpour2015sina}\\
DEANNA~\cite{yahya2012natural}$\ast$ & en & full & 100 & 0.21 & 0.21 & 0.21 & $\approx$ 1-50 s & \cite{zou2014natural}\\
SWIP~\cite{pradel2012semantic} & en & full & 99 & 0.16 & 0.17 & 0.17 & - & \cite{cimiano2013multilingual}\\ 
Zhu et al.~\cite{zhu2015graph}$\ast$ & en & full & 99 & 0.38 & 0.42 & 0.38 & - & \cite{zhu2015graph}\\
\hline
\multicolumn{9}{|c|}{QALD-4}\\
\hline
Xser~\cite{xu2014xser} & en & full & 50 & 0.72 & 0.71 & 0.72 & - & \cite{unger2014question}\\
\textbf{\NAME} & \textbf{en} & \textbf{key} & \textbf{50} & \textbf{0.76} & \textbf{0.40} & \textbf{0.52} & \textbf{0.32s} & -\\
\textbf{\NAME} & \textbf{en} & \textbf{full} & \textbf{50} & \textbf{0.56} & \textbf{0.30} & \textbf{0.39} & \textbf{0.46s} & -\\
gAnswer~\cite{zou2014natural} & en & full & 50 & 0.37 & 0.37 & 0.37 & 0.973 s & \cite{unger2014question}\\
CASIA~\cite{he2014casia} & en & full & 50 & 0.32 & 0.40 & 0.36 & - & \cite{unger2014question}\\
\textbf{\NAME} & \textbf{de} & \textbf{key} & \textbf{50} & \textbf{0.92} & \textbf{0.20} & \textbf{0.33} & \textbf{0.04s} & -\\
\textbf{\NAME} & \textbf{fr} & \textbf{key} & \textbf{50} & \textbf{0.92} & \textbf{0.20} & \textbf{0.33} & \textbf{0.06s} & -\\
\textbf{\NAME} & \textbf{it} & \textbf{key} & \textbf{50} & \textbf{0.92} & \textbf{0.20} & \textbf{0.33} & \textbf{0.04s} & -\\
\textbf{\NAME} & \textbf{es} & \textbf{key} & \textbf{50} & \textbf{0.92} & \textbf{0.20} & \textbf{0.33} & \textbf{0.05s} & -\\
\textbf{\NAME} & \textbf{de} & \textbf{full} & \textbf{50} & \textbf{0.90} & \textbf{0.20} & \textbf{0.32} & \textbf{0.06s} & -\\
\textbf{\NAME} & \textbf{it} & \textbf{full} & \textbf{50} & \textbf{0.92} & \textbf{0.20} & \textbf{0.32} & \textbf{0.16s} & -\\
\textbf{\NAME} & \textbf{es} & \textbf{full} & \textbf{50} & \textbf{0.90} & \textbf{0.20} & \textbf{0.32} & \textbf{0.06s} & -\\
\textbf{\NAME} & \textbf{fr} & \textbf{full} & \textbf{50} & \textbf{0.86} & \textbf{0.18} & \textbf{0.29} & \textbf{0.09s} & -\\
Intui3~\cite{dima2014answering} & en & full & 50 & 0.23 & 0.25 & 0.24 & - & \cite{unger2014question}\\
ISOFT~\cite{park2014isoft} & en & full & 50 & 0.21 & 0.26 & 0.23 & - & \cite{unger2014question}\\
Hakimov~\cite{hakimov2015applying}$\ast$ & en & full & 50 & 0.52 & 0.13 & 0.21 & - & \cite{hakimov2015applying}\\
\hline
\multicolumn{9}{|c|}{QALD-5}\\
\hline
Xser~\cite{xu2014xser} & en & full & 50 & 0.74 & 0.72 & 0.73 & - & \cite{unger2015question}\\
UTQA~\cite{UTQA} & en & full & 50 & - & - & 0.65 & - & \cite{UTQA} \\
UTQA~\cite{UTQA} & es & full & 50 & 0.55 & 0.53 & 0.54 & - & \cite{UTQA}\\
UTQA~\cite{UTQA} & fs & full & 50 & 0.53 & 0.51 & 0.52 & - & \cite{UTQA}\\
\textbf{\NAME} & \textbf{en} & \textbf{full} & \textbf{50} & \textbf{0.56} & \textbf{0.41} & \textbf{0.47} & \textbf{0.62s} & -\\
\textbf{\NAME} & \textbf{en} & \textbf{key} & \textbf{50} & \textbf{0.60} & \textbf{0.27} & \textbf{0.37} & \textbf{0.50s} & -\\
AskNow\cite{dubey2016asknow} & en & full & 50 & 0.32 & 0.34 & 0.33 & & \cite{dubey2016asknow}\\
QAnswer\cite{ruseti2015qanswer} & en & full & 50 & 0.34 & 0.26 & 0.29 & - & \cite{unger2015question}\\
\textbf{\NAME} & \textbf{de} & \textbf{full} & \textbf{50} & \textbf{0.92} & \textbf{0.16} & \textbf{0.28} & \textbf{0.20s} & -\\
\textbf{\NAME} & \textbf{de} & \textbf{key} & \textbf{50} & \textbf{0.90} & \textbf{0.16} & \textbf{0.28} & \textbf{0.19s} & -\\
\textbf{\NAME} & \textbf{fr} & \textbf{full} & \textbf{50} & \textbf{0.90} & \textbf{0.16} & \textbf{0.28} & \textbf{0.19s} & -\\
\textbf{\NAME} & \textbf{fr} & \textbf{key} & \textbf{50} & \textbf{0.90} & \textbf{0.16} & \textbf{0.28} & \textbf{0.18s} & -\\
\textbf{\NAME} & \textbf{it} & \textbf{full} & \textbf{50} & \textbf{0.88} & \textbf{0.18} & \textbf{0.30} & \textbf{0.20s} & -\\
\textbf{\NAME} & \textbf{it} & \textbf{key} & \textbf{50} & \textbf{0.90} & \textbf{0.16} & \textbf{0.28} & \textbf{0.18s} & -\\
\textbf{\NAME} & \textbf{es} & \textbf{full} & \textbf{50} & \textbf{0.88} & \textbf{0.14} & \textbf{0.25} & \textbf{0.20s} & -\\
\textbf{\NAME} & \textbf{es} & \textbf{key} & \textbf{50} & \textbf{0.90} & \textbf{0.14} & \textbf{0.25} & \textbf{0.20s} & -\\
SemGraphQA\cite{beaumont2015semgraphqa} & en & full & 50 & 0.19 & 0.20 & 0.20 & - & \cite{unger2015question}\\
YodaQA\cite{baudivsqald} & en & full & 50 & 0.18 & 0.17 & 0.18 & - & \cite{unger2015question}\\
QuerioDali\cite{lopez2016queriodali}& en & full & 50 & ? & ? & ? & ? & \cite{lopez2016queriodali}\\
\hline
\multicolumn{9}{|c|}{QALD-6}\\
\hline

\textbf{\NAME} & \textbf{en} & \textbf{full} & \textbf{100} & \textbf{0.55} & \textbf{0.34} & \textbf{0.42} & \textbf{1.28s} & -\\

\textbf{\NAME} & \textbf{de} & \textbf{full} & \textbf{100} & \textbf{0.73} & \textbf{0.29} & \textbf{0.41} & \textbf{0.41s} & -\\
\textbf{\NAME} & \textbf{de} & \textbf{key} & \textbf{100} & \textbf{0.85} & \textbf{0.27} & \textbf{0.41} & \textbf{0.30s} & -\\
\textbf{\NAME} & \textbf{en} & \textbf{key} & \textbf{100} & \textbf{0.51} & \textbf{0.30} & \textbf{0.37} & \textbf{1.00s} & -\\
SemGraphQA~\cite{beaumont2015semgraphqa} & en & full & 100 & 0.70 & 0.25 & 0.37 & - & \cite{unger2016question}\\
\textbf{\NAME} & \textbf{fr} & \textbf{key} & \textbf{100} & \textbf{0.78} & \textbf{0.23} & \textbf{0.36} & \textbf{0.34s} & -\\
\textbf{\NAME} & \textbf{fr} & \textbf{full} & \textbf{100} & \textbf{0.57} & \textbf{0.22} & \textbf{0.32} & \textbf{0.46s} & -\\
\textbf{\NAME} & \textbf{es} & \textbf{full} & \textbf{100} & \textbf{0.69} & \textbf{0.19} & \textbf{0.30} & \textbf{0.45s} & -\\
\textbf{\NAME} & \textbf{es} & \textbf{key} & \textbf{100} & \textbf{0.83} & \textbf{0.18} & \textbf{0.30} & \textbf{0.35s} & -\\
\textbf{\NAME} & \textbf{it} & \textbf{key} & \textbf{100} & \textbf{0.75} & \textbf{0.17} & \textbf{0.28} & \textbf{0.34s} & -\\
AMUSE~\cite{hakimov2017amuse} & en & full & 100 &  - & - & 0.26 & - &  \cite{hakimov2017amuse} \\
\textbf{\NAME} & \textbf{it} & \textbf{full} & \textbf{100} & \textbf{0.62} & \textbf{0.15} & \textbf{0.24} & \textbf{0.43s} & -\\
AMUSE~\cite{hakimov2017amuse} & es & full & 100 & - & - & 0.20 & - & \cite{hakimov2017amuse} \\
AMUSE~\cite{hakimov2017amuse} & de & full &100 & - & - & 0.16 & - & \cite{hakimov2017amuse} \\
\hline
\end{tabular}
}
\end{table}

\begin{table}
\scalebox{0.78}{
\begin{tabular}{|c|c|c|c|c|c|c|c|c|}
\hline
QA system & Lang & Type & Total & P & R & F & Runtime & Ref\\
\hline
\multicolumn{9}{|c|}{QALD-7}\\
\hline
\textbf{\NAME} & \textbf{en} & \textbf{full} & \textbf{100} & \textbf{0.25} & \textbf{0.28} & \textbf{0.25} & \textbf{1.24s} & -\\
\textbf{\NAME} & \textbf{fr} & \textbf{key} & \textbf{100} & \textbf{0.18} & \textbf{0.16} & \textbf{0.16} & \textbf{0.32s} & -\\
\textbf{\NAME} & \textbf{en} & \textbf{key} & \textbf{100} & \textbf{0.14} & \textbf{0.16} & \textbf{0.14} & \textbf{0.88s} & -\\
\textbf{\NAME} & \textbf{de} & \textbf{full} & \textbf{100} & \textbf{0.10} & \textbf{0.10} & \textbf{0.10} & \textbf{0.34s} & -\\
\textbf{\NAME} & \textbf{de} & \textbf{key} & \textbf{100} & \textbf{0.12} & \textbf{0.10} & \textbf{0.10} & \textbf{0.28s} & -\\
\textbf{\NAME} & \textbf{fr} & \textbf{full} & \textbf{100} & \textbf{0.12} & \textbf{0.10} & \textbf{0.10} & \textbf{0.42s} & -\\
\textbf{\NAME} & \textbf{it} & \textbf{key} & \textbf{100} & \textbf{0.06} & \textbf{0.06} & \textbf{0.06} & \textbf{0.28s} & -\\
\textbf{\NAME} & \textbf{it} & \textbf{full} & \textbf{100} & \textbf{0.04} & \textbf{0.04} & \textbf{0.04} & \textbf{0.34s} & -\\
\hline
\end{tabular}
}
\caption{This table (left column and upper right column) summarizes the results obtained by the QA systems evaluated with QALD-3 (over DBpedia 3.8), QALD-4 (over DBpedia 3.9), QALD-5 (over DBpedia 2014), QALD-6 (over DBpedia 2015-10), QALD-7 (2016-04). We indicated with \qq{$\ast$}\ the systems that did not participate directly in the challenges, but were evaluated on the same benchmark afterwards. We indicate the average running times of a query for the systems where we found them. Even if the runtime evaluations were executed on different hardware, it still helps to give an idea about the scalability.}
\vspace{-3mm}
\label{tab:QALD Benchmark}
\end{table}

\begin{table}
\scalebox{0.78}{
\begin{tabular}{|c|c|c|c|c|c|c|c|c|c|c|c|}
\hline
 QA System  & Lang & Type & Total & P & R & F & Runtime & Ref \\
\hline
\multicolumn{9}{|c|}{QALD-7 task 4, training dataset}\\
\hline

 \textbf{\NAME} & \textbf{en} & \textbf{full} & \textbf{100} & \textbf{0.37} & \textbf{0.39} & \textbf{0.37} & \textbf{1.68s} & -\\
\textbf{\NAME} & \textbf{en} & \textbf{key} & \textbf{100} & \textbf{0.35} & \textbf{0.38} & \textbf{0.35} & \textbf{0.80s} & -\\
\textbf{\NAME} & \textbf{es} & \textbf{key} & \textbf{100} & \textbf{0.31} & \textbf{0.32} & \textbf{0.31} & \textbf{0.45s} & -\\
Sorokin et al. \cite{sorokinend} & en & full & 100 & - & - & 0.29 & - & \cite{sorokinend}\\
\textbf{\NAME} & \textbf{de} & \textbf{key} & \textbf{100} & \textbf{0.27} & \textbf{0.28} & \textbf{0.27} & \textbf{1.13s} & -\\
\textbf{\NAME} & \textbf{fr} & \textbf{key} & \textbf{100} & \textbf{0.27} & \textbf{0.30} & \textbf{0.27} & \textbf{1.14s} & -\\
\textbf{\NAME} & \textbf{fr} & \textbf{full} & \textbf{100} & \textbf{0.27} & \textbf{0.31} & \textbf{0.27} & \textbf{1.05s} & -\\
\textbf{\NAME} & \textbf{es} & \textbf{full} & \textbf{100} & \textbf{0.24} & \textbf{0.26} & \textbf{0.24} & \textbf{0.65s} & -\\
\textbf{\NAME} & \textbf{de} & \textbf{full} & \textbf{100} & \textbf{0.18} & \textbf{0.20} & \textbf{0.18} & \textbf{0.82s} & -\\
\textbf{\NAME} & \textbf{it} & \textbf{full} & \textbf{100} & \textbf{0.19} & \textbf{0.20} & \textbf{0.18} & \textbf{1.00s} & -\\
\textbf{\NAME} & \textbf{it} & \textbf{key} & \textbf{100} & \textbf{0.17} & \textbf{0.18} & \textbf{0.16} & \textbf{0.44s} & -\\
 \hline
\end{tabular}
}
\caption{The table shows the results of \NAME over the QALD-7 task 4 training dataset. We used Wikidata (dated 2016-11-28).}
\label{tab:wikidata_benchmark}
\vspace{-5mm}

\end{table}

\begin{table}
\scalebox{0.78}{
\begin{tabular}{|c|c|c|c|c|c|c|c|c|}
\hline
 QA System  & Lang & Type & Total & Accuracy & Runtime & Ref \\
\hline
\textbf{\NAME}$^{\ast}$ & \textbf{en} & \textbf{full} & \textbf{21687} &  \textbf{0.571} & \textbf{2.1 s} & \textbf{-}\\
Dai et al.$^{\ast}$ & en & full & 21687 & 0.626 & - & \cite{dai2016cfo} \\
Bordes et al.& en & full  & 21687 &  0.627 & - & \cite{bordes2015large}\\
Yin et al.& en & full & 21687 &  0.683 & - &\cite{yin2016simple} \\
Golub and He& en & full & 21687 &  0.709 & - &  \cite{golub2016character}\\
Lukovnikov et al.& en & full & 21687 & 0.712 & - & \cite{lukovnikov2017neural} \\
\hline
\end{tabular}
}
\caption{This table summarizes the QA systems evaluated over SimpleQuestions. Every system was evaluated over FB2M except the ones marked with ($\ast$) which were evaluated over FB5M.} \label{simplequestions}
\vspace{-5mm}
\end{table}

\begin{table}
\scalebox{0.78}{
\begin{tabular}{|K{2.5cm}|c|c|c|c|c|c|c|c|c|c|}
\hline
 Benchmark  & Lang & Type & Total & P & R & F & Runtime \\
\hline
LC-QuAD & en & full & 5000 & 0.59 & 0.38 & 0.46 & 1.5 s  \\
WDAquaCore0Questions & mixed & mixed & 689 & 0.79 & 0.46 & 0.59 & 1.3 s \\
\hline
\end{tabular}
}
\caption{This table summarizes the results of \NAME over some newly appeared benchmarks.} \label{newbenchmarks}
\vspace{-5mm}
\end{table}

\begin{table}
\scalebox{0.78}{
\begin{tabular}{|K{2.5cm}|c|c|c|c|c|c|c|c|c|c|}
\hline
 Dataset  & Lang & Type & Total & P & R & F & Runtime \\
\hline
 DBpedia & en & full & 100 & 0.55 & 0.34 & 0.42 & 1.37 s \\
 All KBs supported & en & full & 100 & 0.49 & 0.39 & 0.43 & 11.94s\\
\hline
\end{tabular}
}
\caption{Comparison on QALD-6 when querying only DBpedia and multiple KBs at the same time.} \label{multiple}
\vspace{-5mm}
\end{table}

\textbf{SimpleQuestions:} SimpleQuestions contains 108442 questions that can be solved using one triple pattern. We trained our system using the first 100 questions in the training set. The results of our system, together with the state-of-the-art systems are presented in Table~\ref{simplequestions}. 
For this evaluation, we restricted the generated queries with one triple-pattern. 
The system performance is 14\% below the state-of-the-art. 
Note that we achieve this result by considering only 100 of the 75810 questions in the training set, and investing 1 hour of manual work for creating lexicalizations for properties manually. 
Concretely, instead of generating a training dataset with 80.000 questions, which can cost several thousands of euros, we invested 1 hour of manual work with the result of loosing (only) 14\% in accuracy!\\
Note that the SimpleQuestions dataset is highly skewed towards certain properties (it contains 1629 properties, the 20 most frequent properties cover nearly 50\% of the questions). 
Therefore, it is not clear how the other QA systems behave with respect to properties not appearing in the training dataset and with respect to keyword questions. 
Moreover, it is not clear how to port the existing approaches to new languages and it is not possible to adapt them to more difficult questions.
These points are solved using our approach. 
Hence, we provided here, for the first time, a quantitative analysis of the impact of big training data corpora on the quality of a QA system.\\

\textbf{LC-QuAD \& WDAquaCore0Questions:}
Recently, a series of new benchmarks have been published. 
LC-QuAD~\cite{trivedi2017lc} is a benchmark containing 5000 English questions and it concentrates on complex questions. 
WDAquaCore0Questions~\cite{dennisBenchmark} is a benchmark containing 689 questions over multiple languages and addressing mainly Wikidata, generated from the logs of a live running QA system. 
The questions are a mixture of real-world keyword and malformed questions. 
In Table~\ref{newbenchmarks}, we present the first baselines for these benchmarks.\\

\textbf{Multiple KBs:}
The only available benchmark that tackles multiple KBs was presented in QALD-4 task~2. 
The KBs are rather small and perfectly interlinked. 
This is not the case over the considered KBs. 
We therefore evaluated the ability to query multiple KBs differently. 
We run the questions of the QALD-6 benchmark, which was designed for DBpedia, both over DBpedia (only) and over DBpedia, Wikidata, MusicBrainz, DBLP and Freebase. 
Note that, while the original questions have a solution over DBpedia, a good answer could also be found over the other datasets. 
We therefore manually checked whether the answers that were found in other KBs are right (independently from which KB was chosen by the QA system to answer it). 
The results are presented in Table \ref{multiple}. \NAME choose 53 times to answer a question over DBpedia, 39 over Wikidata and the other 8 times over a different KB. Note that we get better results when querying multiple KBs. Globally we get better recall and lower precision which is expected. While scalability is an issue, we are able to pick the right KB to find the answer!\\

Note: We did not tackle the WebQuestions benchmark for the following reasons. 
While it has been shown that WebQuestions can be addressed using non-reified versions of Freebase, this was not the original goal of the benchmark. 
More then 60\% of the QA systems benchmarked over WebQuestions are tailored towards its reefication model. 
There are two important points here. 
First, most KBs in the Semantic Web use binary statements. 
Secondly, in the Semantic Web community, many different reefication models have been developed as described in~\cite{gimenez2017ndfluents}.

\subsubsection{Setting}
All experiments were performed on a virtual machine with 4 core of Intel Xeon E5-2667 v3 3.2GH, 16~GB of RAM and 500~GB of SSD disk. Note that the whole infrastructure was running on this machine, i.e. all indexes and the triple-stores needed to compute the answers (no external service was used). The original data dumps sum up to 336~GB. Note that across all benchmarks we can answer a question in less then 2 seconds except when all KBs are queried at the same time which shows that the algorithm should be parallelized for further optimization.

\section{Provided Services for Multilingual and Multi-KB QA}\label{qanary}
We have presented an algorithm that can be easily ported to new KBs and that can query multiple KBs at the same time. 
In the evaluation section, we have shown that our approach is competitive while offering the advantage of being multilingual and robust to keyword questions. 
Moreover, we have shown that it runs on moderate hardware. 
In this section, we describe how we integrated the approach to an actual service and how we combine it to existing services so that it can be directly used by end-users.

\begin{figure}
\begin{center}
\includegraphics[width=0.5\textwidth]{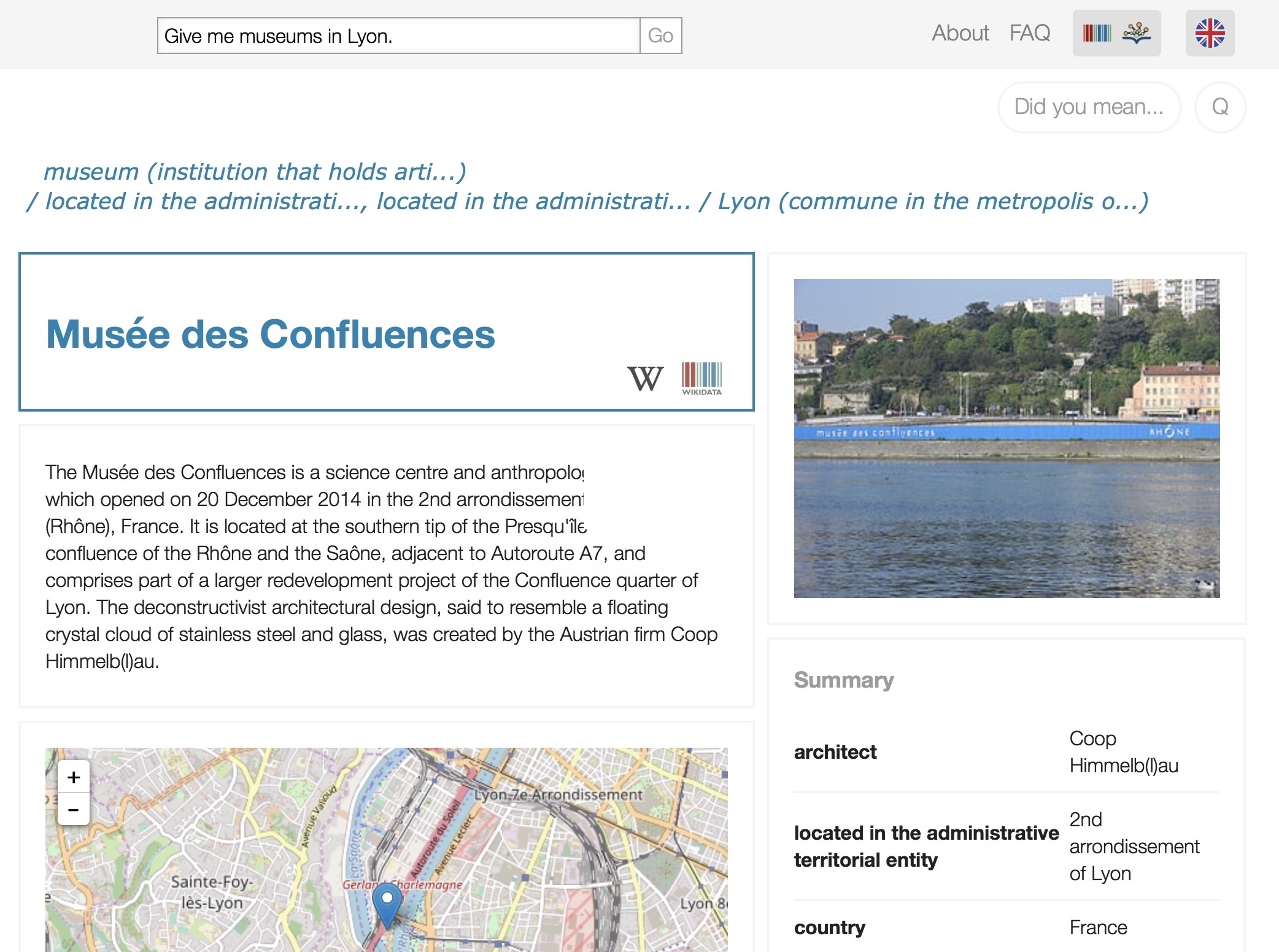}
\caption{Screenshot of Trill, using in the back-end \NAME, for the question \qq{Give me museums in Lyon.}.}\label{fig:screen}     
\end{center}
\vspace{-5mm}
\end{figure}

First, we integrated \NAME\ into Qanary~\cite{bothqanary,dennisqanary}, a framework to integrate QA components. This way \NAME\ can be accessed via RESTful interfaces for example to benchmark it via Gerbil for QA\cite{usbeckabenchmarking}. It also allows to combine it with services that are already integrated into Qanary like a speech recognition component based on Kaldi\footnote{\url{http://kaldi-asr.org}} and a language detection component based on \cite{nakatani2010langdetect}. 
Moreover, the integration into Qanary allows to reuse Trill~\cite{dennisTrill}, a reusable front-end for QA systems. 
A screenshot of Trill using in the back-end \NAME\ can be found in Figure~\ref{fig:screen}.\\
Secondly, we reused and extended Trill to make it easily portable to new KBs.
While Trill originally was supporting only DBpedia and Wikidata, now it supports also MusicBrainz, DBLP and Freebase. 
We designed the extension so that it can be easily ported to new KBs. 
Enabling the support to a new KB is mainly reduced to writing an adapted SPARQL query for the new KB. 
Additionally, the extension allows to select multiple KBs at the same time.\\
Thirdly, we adapted some services that are used in Trill to be easily portable to new KBs.
These include SPARQLToUser~\cite{dennisSparqlToUser}, a tool that generates a human readable version of a SPARQL query and LinkSUM~\cite{linksum} a service for entity summarization. 
All these tools now support the 5 mentioned KBs and the 5 mentioned languages.\\
A public online demo is available under:
\begin{center}
\url{www.wdaqua.eu/qa}
\end{center}


\section{Conclusion and Future Work}\label{conclusion}
In this paper, we introduced a novel concept for QA aimed at multilingual and KB-agnostic QA. 
Due to the described characteristics of our approach portability is ensured which is a significant advantage in comparison to previous approaches.
We have shown the power of our approach in an extensive evaluation over multiple benchmarks.
Hence, we clearly have shown our contributions w.r.t.~qualitative (language, KBs) and quantitative improvements (outperforming many existing systems and querying multiple KBs) as well as the capability of our approach to scale for very large KBs like DBpedia.\\
We have applied our algorithm and adapted a set of existing services so that end-users can query, using multiple languages, multiple KBs at the same time, using an unified interface.
Hence, we provided here a major step towards QA over the Semantic Web following our larger research agenda of providing QA over the LOD cloud.\\
In the future, we want to tackle the following points. 
First, we want to parallelize our approach, s.t.~when querying multiple KBs acceptable response times will be achieved. 
Secondly, we want to query more and more KBs (hints to interesting KBs are welcome). 
Thirdly, from different lessons learned from querying multiple KBs, we want to give a set of recommendations for RDF datasets, s.t.~they are fit for QA. 
And fourth, we want to extend our approach to also query reefied data. 
Fifth, we would like to extend the approach to be able to answer questions including aggregates and functions. 
We believe that our work can further boost the expansion of the Semantic Web since we presented a solution that easily allows to consume RDF data directly by end-users requiring low hardware investments.\\

\textbf{Note: }  There is a Patent Pending for the presented approach. It was submitted the 18 January 2018 at the EPO and has the number EP18305035.0.

\begin{acknowledgements}
This project has received funding from the European Union's Horizon 2020 research and innovation program under the Marie Sklodowska-Curie grant agreement No 642795.
\end{acknowledgements}

\bibliographystyle{ACM-Reference-Format}
\bibliography{bib}

\end{document}